\documentclass{article}
\usepackage{ijcai20}
\usepackage{amssymb}
\usepackage{dsfont}
\usepackage{times}
\usepackage{soul}
\usepackage{url}
\usepackage[hidelinks]{hyperref}
\usepackage[utf8]{inputenc}
\usepackage[small]{caption}
\usepackage{graphicx}
\usepackage{amsmath}
\usepackage{amsthm}
\usepackage{booktabs}
\usepackage{algorithm}
\usepackage{algorithmic}
\usepackage{multirow}

\usepackage{comment}
\title{Unsupervised Domain Adaptation of \\ a Pretrained Cross-Lingual Language Model\thanks{This work was supported by Alibaba Group through Alibaba Innovative Research (AIR) Program.}}

\author{
Juntao Li$^{1,2}$\thanks{This work was done when Juntao Li was an intern at the National University of Singapore.}
\and
Ruidan He$^3$\and
Hai Ye$^{2}$\and
Hwee Tou Ng$^2$ \and
Lidong Bing$^3$  \And
Rui Yan$^1$
\affiliations
$^1$Center for Data Science, Academy for
Advanced Interdisciplinary Studies, Peking University\\
$^2$Department of Computer Science, National University of Singapore\\
$^3$DAMO Academy, Alibaba Group\\
\emails
lijuntao@pku.edu.cn, ruidan.he@alibaba-inc.com, \{yeh, nght\}@comp.nus.edu.sg, \\
l.bing@alibaba-inc.com, ruiyan@pku.edu.cn
}

\begin{document}

\maketitle

 \begin{abstract}
Recent research indicates that pretraining cross-lingual language models on large-scale unlabeled texts yields significant performance improvements over various cross-lingual and low-resource tasks. Through training on one hundred languages and terabytes of texts, cross-lingual language models have proven to be effective in leveraging high-resource languages to enhance low-resource language processing and outperform monolingual models. In this paper, we further investigate the cross-lingual and cross-domain (CLCD) setting when a pretrained cross-lingual language model needs to adapt to new domains. Specifically, we propose a novel unsupervised feature decomposition method that can automatically extract domain-specific features and domain-invariant features from the entangled pretrained cross-lingual representations, given unlabeled raw texts in the source language. Our proposed model leverages mutual information estimation to decompose the representations computed by a cross-lingual model into domain-invariant and domain-specific parts. Experimental results show that our proposed method achieves significant performance improvements over the state-of-the-art pretrained cross-lingual language model in the CLCD setting.
The source code of this paper is publicly available at \url{https://github.com/lijuntaopku/UFD}.

\end{abstract}
 \section{Introduction}
Recent progress in deep learning benefits a variety of NLP tasks and leads to significant performance improvements when large-scale annotated datasets are available.
For high-resource languages, e.g., English, it is feasible for many tasks to collect sufficient labeled data to build deep neural models.
However, for many languages, there might not exist enough data in most cases to make full use of the advances of deep neural models.
As such, various cross-lingual transfer learning methods have been proposed to utilize labeled data from high-resource languages to construct deep models for low-resource languages \cite{kim2019effective,lin2019choosing,he2019cross,vulic2019multilingual}. Nonetheless, most cross-lingual transfer learning research focuses on mitigating the discrimination of languages, while leaving the domain gap less explored.
In this study, we concentrate on a more challenging setting, i.e., cross-lingual and cross-domain (CLCD) transfer, where in-domain labeled data in the source language is not available.

Conventionally, cross-lingual methods mainly rely on extracting language-invariant features from data to transfer knowledge learned from the source language to the target language.
One straightforward method is weight sharing, which directly reuses the model parameters trained on the source language to the target language, by mapping an input text to a shared embedding space beforehand.
However, previous research \cite{chen2018adversarial} revealed that weight sharing is not sufficient for extracting language-invariant features that can generalize well across languages. As a result, a language-adversarial training strategy was proposed to extract invariant features across languages, using non-parallel unlabeled texts from each language.
Such a strategy performs well for the bilingual transfer setting but is not suitable for extracting language-invariant features from multiple languages, since features shared by all source languages might be too sparse to retain useful information.

Recently, pretrained cross-lingual language models at scale, e.g., multilingual BERT \cite{devlin2019bert} and XLM \cite{lample2019cross,conneau2019unsupervised}, show very competitive performance over various cross-lingual tasks, and even outperform pretrained monolingual models on low-resource languages.
Through employing parallel texts (unlabeled for any specific task) and shared sub-word vocabulary over all languages, these pretrained cross-lingual models can effectively encode input texts from multiple languages to one single representation space, which is a feature space shared by multiple languages (more than one hundred).
While generalizing well for extracting language-invariant features, cross-lingual pretraining methods have no specific strategy for extracting domain-invariant features. In our CLCD setting, both language-invariant and domain-invariant features need to be extracted.

To address the aforementioned limitation of cross-lingual pretrained models \cite{conneau2019unsupervised} in CLCD scenarios, we propose an unsupervised
feature decomposition (UFD) method, which only leverages unlabeled data in the source language.
Specifically, our proposed method is inspired by the recently proposed unsupervised representation learning method \cite{hjelm2018learning} and can simultaneously extract domain-invariant features and domain-specific features by combining mutual information maximization and minimization.
Compared to previous cross-lingual transfer learning methods, our proposed model maintains the merits of cross-lingual pretrained models, i.e., generalizing well for over a hundred languages, and only needs unlabeled data in the source language for domain adaptation, which is suitable for more cross-lingual transfer scenarios.

We evaluate our model on a benchmark cross-lingual sentiment classification dataset, i.e., Amazon Review~ \cite{prettenhofer-stein-2010-cross}, which involves multiple languages and domains.
Experimental results indicate that, with the enhancement of the pretrained XLM cross-lingual language model, our proposed UFD model (trained on some unlabeled raw texts in the source language) along with a simple linear classifier (trained on a small labeled dataset in the source language and the source domain) outperforms state-of-the-art models that have access to strong cross-lingual supervision (e.g., commercial MT systems) or labeled datasets in multiple source languages.
Furthermore, incorporating our proposed UFD strategy with an unlabeled set of 150K instances in the source language leads to continuous gains over the strong pretrained XLM model that is trained on one hundred languages and terabytes of texts. 
Extensive experiments further demonstrate that unsupervised feature decomposition on a pretrained cross-lingual language model outperforms a pretrained domain-specific language model trained on over 100 million sentences.

 \section{Related Work}
Cross-lingual transfer learning (CLTL) has long been investigated \cite{yarowsky2001inducing} and is still one of the frontiers of natural language processing \cite{chen2019multi}.
Through utilizing rich annotated data in high-resource languages, CLTL significantly alleviates the challenge of scarce training data in low-resource languages.
Conventionally, CLTL mainly focuses on resources that are available for transferring, e.g., collecting parallel texts between two languages to directly transfer model built in a rich-resource language to a low-resource one \cite{pham2015learning} or constructing annotated data in the target language by machine translation systems \cite{xu2017cross}.
Subsequently, with the success of deep learning, cross-lingual word embeddings are proposed to learn the shared representation space at the fundamental level and can benefit various downstream tasks \cite{artetxe2018robust,conneau2017word}.
Later, a cross-lingual sentence representation is also proposed \cite{conneau2018xnli}.
Chen et al. \shortcite{chen2018adversarial} designed a language-adversarial training strategy to extract language-invariant features that can directly transfer to the target language.

Another direction is pretraining cross-lingual \cite{lample2019cross} or multilingual language models \cite{devlin2019bert}.
Benefiting from the large-scale training texts and model size, these pretraining methods have changed the face of cross-lingual transfer learning.
Empirical results demonstrate that representation space shared by one hundred languages can significantly outperform the language-specific pretrained models \cite{conneau2019unsupervised}.
As language adversarial training will lead to sparse language-invariant representations when multiple languages are involved \cite{chen2019multi}, 
we follow the line of cross-lingual language model pretraining.
Unlike previous pretraining methods, we focus on domain adaptation of these pretrained models.
To maintain the generalization ability of the cross-lingual pretrained model, we mainly consider the unsupervised domain adaptation setting.
The work most related to ours is proposed for unsupervised representation learning \cite{hjelm2018learning}, which is primarily used for visual representation learning.
 \section{Model}
In this section, we first define the problem discussed in this paper and then describe the proposed method in detail. 

\subsection{Problem Definition \& Model Overview}
In this paper, we consider a setting where we only have a labeled set $D_{s,s}$ of a specific language and a specific domain which we call source language and source domain, and we want to train a classifier to be tested on a set $D_{t,t}$ of a different language and a different domain which we call target language and target domain. We also assume access to some unlabeled raw data $D_{s,u}$ of multiple domains including the target domain from the source language during the training phase, which is usually feasible in practical applications. We call this setting unsupervised cross-lingual and cross-domain (CLCD) adaptation. 

\begin{figure}[!t]
\centering
\includegraphics[width=1\columnwidth]{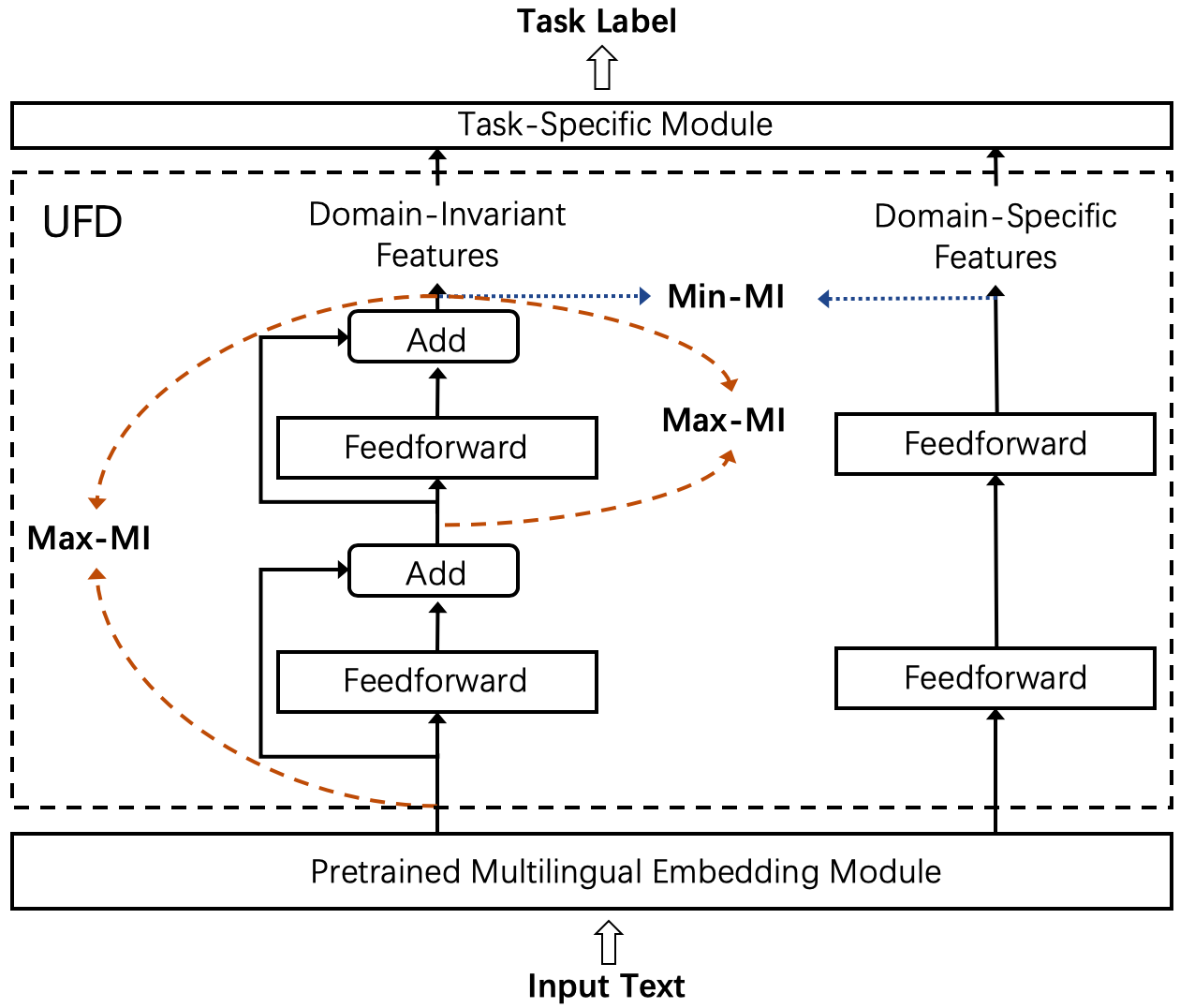}
\caption{\label{fig:mcl}
Our unsupervised domain adaptation model, where Min-MI and Max-MI refer to MI maximization and minimization. The middle-left part is the feature extractor $\mathcal{F}_s$ and the right is $\mathcal{F}_p$. }
\end{figure}

As illustrated in Figure~\ref{fig:mcl}, the proposed method consists of three components: a pretrained multilingual embedding module which embeds the input document into a language-invariant representation, an unsupervised feature decomposition (UFD) module which extracts domain-invariant features and domain-specific features from the entangled language-invariant representation, and a task-specific module trained on the extracted domain-invariant and domain-specific features. 
We adopt XLM\footnote{The latest version XLM-R is adopted, which is trained on over one hundred languages and 2.5 terabytes of texts.}~\cite{lample2019cross} as the multilingual embedding module in our method, which has been pretrained by large-scale parallel and monolingual data from various languages and is the current state-of-the-art cross-lingual language model. We describe the other two modules and the training process in the following subsections.

\subsection{Unsupervised Feature Decomposition}
\textbf{Mutual Information Estimation}

\noindent Before elaborating on the proposed unsupervised feature decomposition module, we first present some preliminary knowledge on mutual information estimation, which is employed in the training objectives of UFD.
Mutual information (MI) is growing in popularity as an objective function in unsupervised representation learning. It measures how informative one variable is of another variable. In the context of unsupervised representation learning, MI maximization is usually adopted such that the encoded representation maximally encodes information of the original data. MI is difficult to compute, particularly in continuous and high-dimensional settings, and therefore various estimation approaches have been proposed. 

In our method, we adopt a recently proposed neural estimation approach \cite{belghazi2018mutual}, which estimates MI of two continuous random variables $X$ and $Y$ by training a network to distinguish between samples coming from their joint distribution, $\mathbb J$, and the product of their marginal distributions, $\mathbb M$. This estimation utilizes a lower-bound of MI based on the Donsker-Varadhan representation (DV) of KL-divergence \cite{donsker1983asymptotic},
\begin{equation}
  \begin{aligned}
    \mathcal{I}(X;Y):= \mathcal{D}_{KL}(\mathbb{J}||\mathbb{M}) \geq \widehat{\mathcal{I}}^{DV}(X;Y)\\
    :=E_\mathbb{J}[T_\omega(x,y)]-\log E_\mathbb{M}[e^{T_\omega(x,y)}]
  \end{aligned}
\end{equation}
where $T_\omega$ is a discrimination function parameterized by a neural network with learnable parameters $\omega$. It maps a sample from space $X \times Y$ to a real value in $\mathbb{R}$. Through maximizing $\widehat{\mathcal{I}}^{DV}$, $T_\omega$ is encouraged to distinguish between samples drawn from $\mathbb J$ and $\mathbb M$ by assigning the former large values while the latter small ones.
\medskip

\noindent\textbf{Proposed Method}

\noindent Let $X \in \mathbb{R}^d$ denote the language-invariant representation generated by the pretrained multilingual embedding module. It is then fed into the proposed UFD module as input. As shown in Figure~\ref{fig:mcl}, we introduce two feature extractors: the domain-invariant extractor $\mathcal{F}_s$ (i.e., the two-layer feedforward network with ReLU activation on the left), and the domain-specific extractor $\mathcal{F}_p$ (i.e., the two-layer network on the right). We denote the extracted features as $\mathcal{F}_s(X)$ and $\mathcal{F}_p(X)$ respectively. Note that for $\mathcal{F}_s$, we add residual connections to better maintain domain-invariant attributes from $X$.

Specifically, $\mathcal{F}_s$ aims to extract domain-invariant features from the language-invariant representation in an unsupervised manner. Since the multilingual embedding module is pretrained on open domain datasets from over one hundred languages, presumably the generated language-invariant representations should contain certain attributes that can be generalized across domains. When $\mathcal{F}_s$ is trained on multiple domains with jointly maximizing MI between the inputs and outputs of each domain, it is encouraged to retain the shared features among those domains from the language-invariant representations. In this way, $\mathcal{F}_s$ is forced to pass domain-invariant information from $X$ to $\mathcal{F}_s(X)$. 

We utilize the neural network-based estimator as presented in Equation (1) for computing MI. In our case, as $\mathcal{F}_s(X)$ is dependent on $X$, we can simplify the DV-based MI estimator to a Jensen-Shannon MI estimator as suggested in~\cite{hjelm2018learning}:
\begin{equation} \label{eq:mi}
  \begin{aligned}
\widehat{\mathcal{I}}^{JSD}(X;\mathcal{F}_s(X)) :=E_{\mathbb{P}}[-sp(-T_\omega(x,\mathcal{F}_s(x)))]\\
-E_{\mathbb{P}\times\widetilde{\mathbb{P}}}[sp(T_\omega(x^{'},\mathcal{F}_s(x)))]
  \end{aligned}
\end{equation}
where $x$ is an input embedding with empirical probability distribution $\mathbb P$. As $\mathcal{F}_s(x)$ is directly computed from $x$, $(x,\mathcal{F}_s(x))$ can be regarded as a sample drawn from the joint distribution of $X$ and $\mathcal{F}_s(X)$. $x^{'}$ corresponds to an input embedding from $\widetilde{\mathbb{P}}=\mathbb{P}$, i.e., $x^{'}$ is computed from a random sample drawn from the same input distribution, such that $(x^{'}, \mathcal{F}_s(x))$ is drawn from the product of marginal distributions.
$sp(z)=\log (1+e^z)$ is the softplus activation function. The training objective of $\mathcal{F}_s$ is to maximize the MI on $X$ and $\mathcal{F}_s(X)$ and the loss is formulated as follows:
\begin{equation}
  \begin{aligned}
\mathcal{L}_s(\omega_s, \psi_s)  = - \widehat{\mathcal{I}}^{JSD}(X, \mathcal{F}_s(X))
  \end{aligned}
\end{equation}
where $\omega_s$ denotes the parameters of the discrimination network in the estimator and $\psi_s$ denotes the parameters of $\mathcal{F}_s$.
To facilitate learning of domain-invariant features, we also propose to maximize the MI on $\mathcal{F}_s(X)$ and the corresponding intermediate representation (first layer output) $\mathcal{F}_s^{\prime}(X)$, and the training loss is as follows:
\begin{equation}
  \begin{aligned}
\mathcal{L}_r(\omega_r, \psi_s)  = - \widehat{\mathcal{I}}^{JSD}(\mathcal{F}_s^{\prime}(X), \mathcal{F}_s(X))
  \end{aligned}
\end{equation}
where $\omega_r$ denotes the parameters of the discriminator network in the estimator.

Recall that the objective of $\mathcal{F}_p$ is to extract domain-specific features, which is supposed to be exclusive and independent of domain-invariant features.
We propose to minimize the MI between features extracted by $\mathcal{F}_s$ and $\mathcal{F}_p$, and the training loss is formulated as follows:
\begin{equation}
  \begin{aligned}
\mathcal{L}_p(\omega_p, \psi_s, \psi_p)  = \widehat{\mathcal{I}}^{JSD}(\mathcal{F}_s(X), \mathcal{F}_p(X))
  \end{aligned}
\end{equation}
where $\psi_p$ denotes the parameters of $\mathcal{F}_p$. $\omega_p$ denotes the parameters of the discrimination network in MI estimator.

The training objective of the proposed UFD component is thus to minimize the overall loss as follows:
\begin{equation}
  \begin{aligned}
\mathcal{L}_{UFD} = \alpha\mathcal{L}_s + \beta\mathcal{L}_r +\gamma\mathcal{L}_p
  \end{aligned}
\end{equation}
where $\alpha$, $\beta$, and $\gamma$ are hyperparameters to balance the effects of sub-losses.

\subsection{Task-Specific Module}
In the task-specific module, we first employ a linear layer that maps the concatenation of the domain-invariant and domain-specific features in $\mathbb{R}^{2d}$ into a vector representation in $\mathbb{R}^{d}$. A simple feedforward layer with softmax activation is then employed on this mapped vector representation to output the task label. We train this module on $D_{s,s}$ and the cross-entropy loss denoted as $\mathcal{L}_t$ is utilized as the training objective.

\subsection{Training}
Note that the parameters of the multilingual embedding module are pretrained and set to be frozen in the entire training process. 
We first optimize the parameters of UFD, i.e., \{$\widehat{\omega}_s,\widehat{\omega}_r,\widehat{\omega}_p, \widehat{\psi}_s,  \widehat{\psi}_p$\} by minimizing $\mathcal{L}_{UFD}$ on the unlabeled set $D_{s,u}$.
Once the UFD module is trained, we fix its parameters and train the task-specific module by minimizing $\mathcal{L}_t$ on the labeled set $D_{s,s}$.
 \section{Experimental Setting}
\begin{table}[t]
\begin{center}
\footnotesize
\begin{tabular}{|l|| r| r| r|}
\hline
\multirow{2}*{\bf Datasets } &\multicolumn{3}{c|}{\bf English} \\
\cline{2-4}
& \bf~Books~ & \bf ~DVD~ &\bf ~Music~ \\
\hline
\hline
\#Documents & 8,898,041 & 1,097,592 & 1,697,533\\
\#Sentences & 101,061,948  & 16,447,191 &21,062,292 \\
\#Words &1,302,754,313 & 194,145,510 & 277,987,802\\
Avg length &146.4 & 176.9 & 163.8 \\
\hline
\end{tabular}
\caption{\label{tab:domain_raw} Statistics of domain-specific raw texts.}
\end{center}
\end{table}

\subsection{Datasets}
We conduct experiments on the multi-lingual and multi-domain Amazon review dataset \cite{prettenhofer-stein-2010-cross}, which serves as a benchmark in previous cross-lingual sentiment analysis research and also supports cross-lingual and cross-domain evaluation.
This dataset includes texts in four languages, i.e., English, German, French, and Japanese, and each language contains three domains, i.e., Books, DVD, and Music. 
There are a training set and a test set for each domain in each language and both consist of 1,000 positive reviews and 1,000 negative reviews.

In our CLCD evaluation, we treat English as the only source language and attempt to adapt to the other three languages. As each language contains three domains, we can construct $3 \times 2$ CLCD source-target pairs between English and a specific target language. Therefore, we have 18 CLCD source-target pairs in total considering all three target languages. During training, we first utilize some unlabeled raw data from the source language for optimizing the proposed UFD. Then, the training set from the source language and source domain is used for training the task-specific module. During testing, the model is evaluated on the test set of the target language and target domain. 

We draw samples from 3 larger unannotated datasets of Books, DVD, and Music domains released in \cite{he2016ups}. The statistics of the three datasets are given in Table \ref{tab:domain_raw}. We randomly sample 50K documents from each domain as the unlabeled domain-specific set in the source language (i.e., English) to be utilized during training.
To encourage $\mathcal{F}_s$ to capture domain-shared features, we utilize domain-specific unlabeled sets from all domains (50K $\times$ 3) in training the UFD module.
We also show the change in performance when varying the number of unlabeled samples in Section 5.

\subsection{Baselines}

We denote our proposed model as \textbf{XLM-UFD}, and we compare it with the following baselines:

\begin{table*}[t]
\begin{center}
\footnotesize
\resizebox{\textwidth}{!}{
\begin{tabular}{|l||c c c c||c c c c|| c c c c |}
\hline
\multirow{2}*{\bf Model } & \multicolumn{4}{c||}{\bf German} & \multicolumn{4}{c||}{\bf French} & \multicolumn{4}{c|}{\bf Japanese}\\
\cline{2-13}
& \bf~Books~ & \bf ~DVD~ &\bf ~Music~ & \bf ~Avg~ & \bf~Books~ & \bf ~DVD~ &\bf ~Music~ & \bf ~Avg~ & \bf~Books~ & \bf ~DVD~ &\bf ~Music~ & \bf ~Avg~\\
\hline
\hline
CL-RL & 79.9  & 77.1 & 77.3 & 78.1 &78.3 &  74.8 & 78.7 &77.3 & 71.1 & 73.1 & 74.4 &72.9\\
Bi-PV & 79.5  & 78.6 & 82.5 & 80.2 &84.3 &  79.6 & 80.1 &81.3 & 71.8 & 75.4 & 75.5 &74.2\\
CLDFA &84.0 &83.1 &  79.0 & 82.0 & 83.4  &82.6 & 83.3 & 83.1 & 77.4  &80.5 & 76.5 & 78.1\\
MAN-MOE &82.4 & 78.8 & 77.2 &79.5  &81.1 &  84.3 & 80.9 &82.1& 62.8  &69.1 & 72.6 &68.2\\
\hline
ADAN & 82.7&	77.1&	79.2&	79.6&		75.9&	75.2&	73.8&	74.9&	72.5&	72.3&	74.3&	73.0\\
MAN-MOE-D &82.8&	80.1&	81.6&	81.5&	83.0&	85.5&	82.0&	83.5&	70.5&	76.0&	70.8&	72.4\\
Multi-BPE & 51.0 & 53.4 & 53.0 & 52.5 &50.5& 51.4 & 51.1& 51.0&50.0&49.8&50.0&49.9\\
DLM &52.1&	53.7&	53.3&	53.0& 57.4&	51.5&	55.2&	54.7&	52.8&	51.5&	50.8&	51.7\\
XLM &80.4&	84.9&	79.3&	81.5&	86.4&	86.3&	83.2&	85.3&	81.7&	81.6&	84.1&	82.5\\
XLM-UFD &\bf 89.2&\bf 86.4&\bf	88.8&\bf	88.1&\bf		89.5&	\bf 89.4&\bf	89.1&	\bf89.3&\bf	83.8&\bf	84.5&\bf	85.2&\bf	84.5\\
\hline
XLM* & 86.3&	81.2&	84.5&	84.0&	90.6&	86.9&	87.6&	88.4&	82.9&	85.0&	87.0&	85.0\\
\hline
\end{tabular}}
\caption{\label{tab:overall} Overall comparison of classification accuracy between our proposed model and baseline models. The upper part refers to the accuracy reported in previous studies in a cross-lingual setting while the middle part refers to our implemented models trained in a CLCD setting. 
XLM* denotes the XLM model trained on source language target domain labeled data. We report the average values of three runs.
}
\end{center}
\end{table*}

\textbf{CL-RL} \cite{xiao2013semi} is a cross-lingual word representation learning method, which learns the connection between two languages by sharing part of the word vectors.

\textbf{Bi-PV} \cite{pham2015learning} attempts to learn paragraph vectors in a bilingual context setting by sharing the distributed representations of unannotated parallel data from different languages.

\textbf{CLDFA} \cite{xu2017cross} is a cross-lingual distillation method which leverages a parallel corpus of documents. An adversarial feature adaptation strategy is applied for reducing the mismatch between the labeled data and the unlabeled parallel document.

\textbf{MAN-MOE} \cite{chen2019multi} addresses the multi-lingual transfer setting, i.e., there are multiple source languages with labeled data.
Building upon a language-adversarial training module, this model utilizes a mixture-of-experts (MOE) module to dynamically combine private features of different languages. 

The above four baselines were originally proposed for adaptation in a cross-lingual setting, e.g., adapting from English-Books to German-Books. We report their official results released in the original papers, which can be regarded as upper bounds for their CLCD performances. Note that the setting of MAN-MOE is different, where N to 1 adaption is performed, i.e., from N source languages to one target language. Thus, its cross-lingual performance cannot be simply viewed as the upper bound of its CLCD performance. We retrain the model in the CLCD setting as another baseline described later. For the baselines described below, they are all trained in the CLCD setting.

\textbf{ADAN} \cite{chen2018adversarial} exploits adversarial training to reduce the representation discrepancy between the encoded source and target embeddings. 

\textbf{MAN-MOE-D} is the version of MAN-MOE trained in a CLCD setting. As this specific model performs N to 1 adaptation, it can adapt from multiple source domains from the same source language to a specific target domain and target language. In our experiments, MAN-MOE-D utilizes two source domains from the same source language. For example, when the target language and domain are German-Books, MAN-MOE-D takes labeled set from both English-DVD and English-Music during training.

\textbf{Multi-BPE} combines the pretrained multilingual byte-pair embeddings in 275 languages \cite{heinzerling2018bpemb}\footnote{https://nlp.h-its.org/bpemb/multi/}  with the task-specific  classifier used in our proposed model to perform CLCD adaptation. This model is used to calibrate the performance of the subword embeddings shared across multiple languages.

\textbf{DLM} is a pretrained domain-specific language model\footnote{Trained with the datasets presented in Table 1.} implemented with the code of XLM\footnote{https://github.com/facebookresearch/XLM} \cite{lample2019cross}.  It employs the pretrained multilingual byte-pair embeddings as the initialized representations of input texts to mitigate the gap between the source language and the target language. This model is used to study the effect of leveraging large scale domain-specific unlabeled texts.

\textbf{XLM} refers to the model where we simply add a feedforward layer with softmax activation as the output layer on top of pretrained XLM \cite{conneau2019unsupervised}.

\subsection{Training Details}
The hidden dimension of XLM is 1024.
The input and output dimensions of the feedforward layers in both $\mathcal{F}_s$ and $\mathcal{F}_p$ are 1024.
The discriminator of $T_{\omega_s}$, $T_{\omega_r}$, and $T_{\omega_p}$ share the same model structure as suggested in previous work \cite{hjelm2018learning}, i.e., the discriminator consists of two feedforward layers with ReLU activation.
The input and output dimensions of the first feedforward layer in the discriminator are 2048 and 1024.
The input and output dimensions of the second feedforward layer are 1024 and 1.
The input dimension of the single-layer task-specific classifiers is 1024.
All trainable parameters are initialized from a uniform distribution $[-0.1, 0.1]$.

We utilize 100 labeled data in the target language and target domain as the validation set, which is used for hyperparameter tuning and model selection during training. The hyperparameters are tuned on the validation set of a specific source-target pair, and are then fixed in all experiments of XLM-UFD.
Specifically, both UFD and the task-specific module are optimized by Adam \cite{kingma2014adam} with a learning rate of $1 \times 10^{-4}$.
The batch size of training UFD and the task-specific module are set to 16 and 8, respectively. The weights $\alpha$, $\beta$, $\gamma$ in Equation (6) are set to 1, 0.3, and 1, respectively. During training, the model that achieves the best performance (lowest loss) on the validation set is saved for evaluation purpose.
 \section{Results}
\begin{table*}[t]
\begin{center}
\footnotesize
\resizebox{\textwidth}{!}{
\begin{tabular}{|l|l||c c c c||c c c c|| c c c c |}
\hline
\multirow{2}*{} &\multirow{2}*{\bf Settings } & \multicolumn{4}{c||}{\bf German} & \multicolumn{4}{c||}{\bf French} & \multicolumn{4}{c|}{\bf Japanese}\\
\cline{3-14}
& & \bf~Books~ & \bf ~DVD~ &\bf ~Music~ & \bf ~Avg~ & \bf~Books~ & \bf ~DVD~ &\bf ~Music~ & \bf ~Avg~ & \bf~Books~ & \bf ~DVD~ &\bf ~Music~ & \bf ~Avg~\\
\hline
\hline
Basic Model&XLM-MI &80.4&	84.9&	79.3&	81.5&	86.4&	86.3&	83.2&	85.3&	81.7&	81.6&	84.1&	82.5\\
\hline
\hline
\multirow{3}*{Model Ablation}& Max-MI &84.5&	82.0&	81.9&	82.8&	81.1& 83.2&	81.8&	82.0&	81.8&	80.6&	81.5&	81.3\\
&Max-Min-MI &88.4&	85.9&	87.3&	87.2&	88.0&	88.4&	88.3&	88.2&\bf 84.4&	83.3&	85.0&	84.2\\
&2Max-Min-MI & \bf 89.2&\bf 86.4&\bf	88.8&\bf	88.1&\bf		89.5&	\bf 89.4&\bf	89.1&	\bf89.3&	83.8&\bf	84.5&\bf	85.2&\bf	84.5\\
\hline
\hline
\multirow{5}*{Unlabeled Data Size}& 1K$\times$3 &87.2& 85.4&	86.4&	86.4&	88.6&	88.3&	87.0&	88.0&	78.9&	80.0 & 81.0&	80.0\\
&2K$\times$3& 86.7&	84.4&	85.6&	85.6&	87.9&	88.1&	83.8&	86.6&\bf	84.2&	83.4&	84.4&	84.0\\
&5K$\times$3& 89.0&	86.0& 86.9& 87.4&	87.8&	89.1&	86.9&	87.9&	83.0&	83.8&	82.2&	82.9\\
&10K$\times$3 & 88.5&	86.3& 88.1&	87.6&	88.8&	88.8&	88.3&	88.6&	83.7&	84.4&\bf	85.5&\bf 84.5\\
&50K$\times$3 &\bf 89.2&\bf 86.4&\bf	88.8&\bf	88.1&	\bf	89.5&\bf	89.4&\bf	89.1&\bf	89.3&	83.8&\bf	84.5&	85.2&\bf	84.5\\
\hline
\end{tabular}}
\caption{\label{tab:ablation} Classification accuracy of an ablation study and using different sizes of unlabeled data in the source language (i.e., English).}
\end{center}
\end{table*}

Table \ref{tab:overall} presents the model comparison results and Table \ref{tab:ablation} shows the results of different ablation tests on XLM-UFD. Classification accuracy is used as the evaluation metric. 

\subsection{Model Comparison}
In Table 2, the top 4 models are trained in a cross-lingual setting, and the middle 6 models are trained in a CLCD setting. We repeat the experiment on each source-target pair for 3 times with different random seeds and record the average result on each pair. Each reported result for models trained in the CLCD setting is the average result of the adaptation performance from two source domains in English. For example, a result under German-Books is the average of adaptation accuracies from English-DVD and English-Music.

We make the following observations from Table 2. (1) XLM-UFD achieves significantly better results over all baselines across all settings. It even substantially outperforms baselines trained in a cross-lingual setting with parallel texts from source and target languages such as CLDFA, which is a much less challenging setting. (2) One interesting finding is that MAN-MOE-D performs better than MAN-MOE. One possible reason is that MAN-MOE involves multiple source languages while invariant features shared by multiple languages might be too sparse to maintain enough information for extracting task-specific features. (3) Among the pretrained models, multilingual byte pair embeddings (Multi-BPE) only achieves low performance. With the enhancement of large-scale domain-specific unlabeled text, the domain-specific language model (DLM) taking the multilingual byte pair embeddings as input obtains observable performance gains but still has much room for improvement.
Benefiting from the large-scale training data and network size, XLM is able to perform better than the state-of-the-art task-specific models such as CLDFA and MAN-MOE-D on French and Japanese.
When combined with the proposed UFD, significant performance gains are observed on XLM.
This points out that domain adaptation is necessary for pretrained multilingual language models when applied to a specific task. 

\subsection{Ablation Study}
To determine the effect of each module of XLM-UFD, we conduct a thorough model ablation.
As presented in Table \ref{tab:ablation}, we first examine the domain-invariant feature extractor along with MI maximization between the language-invariant features from the multilingual embedding module and the extracted domain-invariant features, namely Max-MI. 
Classification accuracy shows that Max-MI with only domain-invariant features enhances the performance of XLM on German but leads to decreased performance on French and Japanese.
Through supplementing the domain-specific feature extractor and the Min-MI objective (i.e., $\mathcal{L}_p$), Max-Min-MI has a noticeable performance increase over Max-MI and outperforms XLM, which confirms that unsupervised feature decomposition can support dynamic domain-specific and domain-invariant feature combination and improve the task performance. 
With the enhancement of the intermediate Max-MI objective (i.e., $\mathcal{L}_r$) between the intermediate features and the output of domain-invariant feature extractor, 2Max-Min-MI achieves significant performance improvement over Max-MI, and it is used as the full model for conducting other comparison and ablation.
In addition, we investigate the effect of unlabeled data size of the source language used during training. 
It can be seen from Table \ref{tab:ablation} that the setting with 5K$\times$3 unlabeled raw texts already yields a very promising performance.
Further increasing the number of unlabeled examples continuously improves the model performance on French and German.
When the unlabeled data size is larger than 10K$\times$3, the performance improvement becomes marginal on Japanese but continues on German and French.

\begin{figure}[!t]
\centering
\includegraphics[width=1\columnwidth]{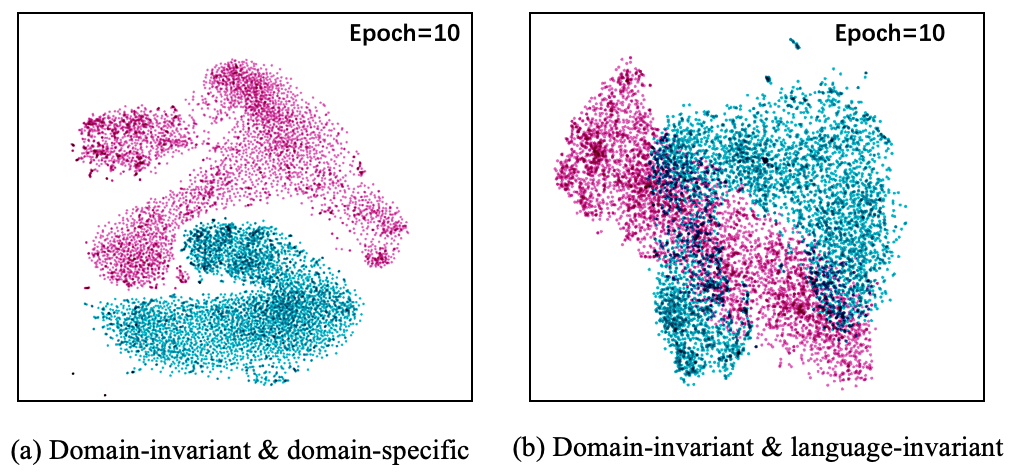}
\caption{\label{fig:visual}
t-SNE plots, where the left figure refers to domain-invariant features and domain-specific features of input texts, and the right figure corresponds to domain-invariant features of input texts and language-invariant representations from XLM.}
\end{figure}

\subsection{Visualization}
To intuitively understand the process of domain-invariant feature and domain-specific feature extraction, we also give the t-SNE plots \cite{maaten2008visualizing} of the UFD module at the tenth epoch. 
Specifically, we sample five thousand raw texts from the source domain and target language.
Each raw text is processed by XLM, and the following domain-invariant feature extractor and domain-specific feature extractor, respectively. 
As presented in Figure \ref{fig:visual}, each data point in the plots represents an input text.
We can observe from the left plot that the domain-invariant features and domain-specific features of input texts have a clear border that can be distinguished, which suggests that mutual information minimization can force the two extractors to exclusively extract two sets of features.
The right plot in Figure \ref{fig:visual} demonstrates that the domain-invariant features and the language-invariant representations from XLM are partly entangled, which can be explained by the fact that maximizing mutual information between them can force the domain-invariant extractor to retain useful features from the language-invariant representations that are shared among different domains. 
 \section{Conclusions and Future Work}
In this paper, we propose a simple but effective unsupervised feature decomposition module that extends the pretrained cross-lingual model to a more useful CLCD scenario. 
Through introducing the mutual information maximization and minimization objectives in representation learning, our proposed method can automatically extract domain-invariant and domain-specific features from the language-invariant cross-lingual space, by using only a small unlabeled dataset from the source language during training.
Experimental results indicate that, with the enhancement of the proposed module, the cross-lingual language model XLM achieves continuous improvements, which leads to new state-of-the-art results on the Amazon review benchmark dataset in a CLCD setting. 
In the future, we will explore the effect of our proposed unsupervised feature decomposition model on other pretrained models and downstream tasks.

\section*{Acknowledgments}
The authors gratefully acknowledge the assistance of Qingyu Tan in this work.

 \bibliographystyle{named}
 \bibliography{ijcai20}

\begin{thebibliography}{}

\bibitem[\protect\citeauthoryear{Artetxe \bgroup \em et al.\egroup
  }{2018}]{artetxe2018robust}
Mikel Artetxe, Gorka Labaka, and Eneko Agirre.
\newblock A robust self-learning method for fully unsupervised cross-lingual
  mappings of word embeddings.
\newblock In {\em ACL}, pages 789--798, 2018.

\bibitem[\protect\citeauthoryear{Belghazi \bgroup \em et al.\egroup
  }{2018}]{belghazi2018mutual}
Mohamed~Ishmael Belghazi, Aristide Baratin, Sai Rajeshwar, Sherjil Ozair,
  Yoshua Bengio, Devon Hjelm, and Aaron Courville.
\newblock Mutual information neural estimation.
\newblock In {\em ICML}, pages 530--539, 2018.

\bibitem[\protect\citeauthoryear{Chen \bgroup \em et al.\egroup
  }{2018}]{chen2018adversarial}
Xilun Chen, Yu~Sun, Ben Athiwaratkun, Claire Cardie, and Kilian Weinberger.
\newblock Adversarial deep averaging networks for cross-lingual sentiment
  classification.
\newblock {\em TACL}, 6:557--570, 2018.

\bibitem[\protect\citeauthoryear{Chen \bgroup \em et al.\egroup
  }{2019}]{chen2019multi}
Xilun Chen, Ahmed Hassan~Awadallah, Hany Hassan, Wei Wang, and Claire Cardie.
\newblock Multi-source cross-lingual model transfer: Learning what to share.
\newblock In {\em ACL}, 2019.

\bibitem[\protect\citeauthoryear{Conneau and Lample}{2019}]{lample2019cross}
Alexis Conneau and Guillaume Lample.
\newblock Cross-lingual language model pretraining.
\newblock In {\em NeurIPS}, 2019.

\bibitem[\protect\citeauthoryear{Conneau \bgroup \em et al.\egroup
  }{2018a}]{conneau2017word}
Alexis Conneau, Guillaume Lample, Marc'Aurelio Ranzato, Ludovic Denoyer, and
  Herv{\'e} J{\'e}gou.
\newblock Word translation without parallel data.
\newblock In {\em ICLR}, 2018.

\bibitem[\protect\citeauthoryear{Conneau \bgroup \em et al.\egroup
  }{2018b}]{conneau2018xnli}
Alexis Conneau, Ruty Rinott, Guillaume Lample, Adina Williams, Samuel Bowman,
  Holger Schwenk, and Veselin Stoyanov.
\newblock Xnli: Evaluating cross-lingual sentence representations.
\newblock In {\em EMNLP}, pages 2475--2485, 2018.

\bibitem[\protect\citeauthoryear{Conneau \bgroup \em et al.\egroup
  }{2019}]{conneau2019unsupervised}
Alexis Conneau, Kartikay Khandelwal, Naman Goyal, Vishrav Chaudhary, Guillaume
  Wenzek, Francisco Guzm{\'a}n, Edouard Grave, Myle Ott, Luke Zettlemoyer, and
  Veselin Stoyanov.
\newblock Unsupervised cross-lingual representation learning at scale.
\newblock {\em arXiv preprint arXiv:1911.02116}, 2019.

\bibitem[\protect\citeauthoryear{Devlin \bgroup \em et al.\egroup
  }{2019}]{devlin2019bert}
Jacob Devlin, Ming-Wei Chang, Kenton Lee, and Kristina Toutanova.
\newblock Bert: Pre-training of deep bidirectional transformers for language
  understanding.
\newblock In {\em ACL}, pages 4171--4186, 2019.

\bibitem[\protect\citeauthoryear{Donsker and
  Varadhan}{1983}]{donsker1983asymptotic}
Monroe~D Donsker and SR~Srinivasa Varadhan.
\newblock Asymptotic evaluation of certain markov process expectations for
  large time. {IV}.
\newblock {\em Communications on Pure and Applied Mathematics}, 36(2):183--212,
  1983.

\bibitem[\protect\citeauthoryear{He and McAuley}{2016}]{he2016ups}
Ruining He and Julian McAuley.
\newblock Ups and downs: Modeling the visual evolution of fashion trends with
  one-class collaborative filtering.
\newblock In {\em WWW}, pages 507--517, 2016.

\bibitem[\protect\citeauthoryear{He \bgroup \em et al.\egroup
  }{2019}]{he2019cross}
Junxian He, Zhisong Zhang, Taylor Berg-Kiripatrick, and Graham Neubig.
\newblock Cross-lingual syntactic transfer through unsupervised adaptation of
  invertible projections.
\newblock {\em arXiv preprint arXiv:1906.02656}, 2019.

\bibitem[\protect\citeauthoryear{Heinzerling and
  Strube}{2018}]{heinzerling2018bpemb}
Benjamin Heinzerling and Michael Strube.
\newblock Bpemb: Tokenization-free pre-trained subword embeddings in 275
  languages.
\newblock In {\em LREC}, 2018.

\bibitem[\protect\citeauthoryear{Hjelm \bgroup \em et al.\egroup
  }{2019}]{hjelm2018learning}
R~Devon Hjelm, Alex Fedorov, Samuel Lavoie-Marchildon, Karan Grewal, Phil
  Bachman, Adam Trischler, and Yoshua Bengio.
\newblock Learning deep representations by mutual information estimation and
  maximization.
\newblock In {\em ICLR}, 2019.

\bibitem[\protect\citeauthoryear{Kim \bgroup \em et al.\egroup
  }{2019}]{kim2019effective}
Yunsu Kim, Yingbo Gao, and Hermann Ney.
\newblock Effective cross-lingual transfer of neural machine translation models
  without shared vocabularies.
\newblock {\em arXiv preprint arXiv:1905.05475}, 2019.

\bibitem[\protect\citeauthoryear{Kingma and Ba}{2014}]{kingma2014adam}
Diederik~P Kingma and Jimmy Ba.
\newblock Adam: A method for stochastic optimization.
\newblock {\em arXiv preprint arXiv:1412.6980}, 2014.

\bibitem[\protect\citeauthoryear{Lin \bgroup \em et al.\egroup
  }{2019}]{lin2019choosing}
Yu-Hsiang Lin, Chian-Yu Chen, Jean Lee, Zirui Li, Yuyan Zhang, Mengzhou Xia,
  Shruti Rijhwani, Junxian He, Zhisong Zhang, Xuezhe Ma, et~al.
\newblock Choosing transfer languages for cross-lingual learning.
\newblock {\em arXiv preprint arXiv:1905.12688}, 2019.

\bibitem[\protect\citeauthoryear{Maaten and
  Hinton}{2008}]{maaten2008visualizing}
Laurens van~der Maaten and Geoffrey Hinton.
\newblock Visualizing data using t-sne.
\newblock {\em JMLR}, 9(Nov):2579--2605, 2008.

\bibitem[\protect\citeauthoryear{Pham \bgroup \em et al.\egroup
  }{2015}]{pham2015learning}
Hieu Pham, Thang Luong, and Christopher Manning.
\newblock Learning distributed representations for multilingual text sequences.
\newblock In {\em Proceedings of the 1st Workshop on Vector Space Modeling for
  Natural Language Processing}, pages 88--94, 2015.

\bibitem[\protect\citeauthoryear{Prettenhofer and
  Stein}{2010}]{prettenhofer-stein-2010-cross}
Peter Prettenhofer and Benno Stein.
\newblock Cross-language text classification using structural correspondence
  learning.
\newblock In {\em ACL}, pages 1118--1127, 2010.

\bibitem[\protect\citeauthoryear{Vuli{\'c} \bgroup \em et al.\egroup
  }{2019}]{vulic2019multilingual}
Ivan Vuli{\'c}, Simone~Paolo Ponzetto, and Goran Glava{\v{s}}.
\newblock Multilingual and cross-lingual graded lexical entailment.
\newblock In {\em ACL}, pages 4963--4974, 2019.

\bibitem[\protect\citeauthoryear{Xiao and Guo}{2013}]{xiao2013semi}
Min Xiao and Yuhong Guo.
\newblock Semi-supervised representation learning for cross-lingual text
  classification.
\newblock In {\em EMNLP}, pages 1465--1475, 2013.

\bibitem[\protect\citeauthoryear{Xu and Yang}{2017}]{xu2017cross}
Ruochen Xu and Yiming Yang.
\newblock Cross-lingual distillation for text classification.
\newblock In {\em ACL}, pages 1415--1425, 2017.

\bibitem[\protect\citeauthoryear{Yarowsky \bgroup \em et al.\egroup
  }{2001}]{yarowsky2001inducing}
David Yarowsky, Grace Ngai, and Richard Wicentowski.
\newblock Inducing multilingual text analysis tools via robust projection
  across aligned corpora.
\newblock In {\em HLT}, pages 1--8, 2001.

\end{thebibliography}

\appendix

\end{document}